\documentclass[lettersize,journal]{IEEEtran}
\usepackage{amsmath,amsfonts}
\usepackage{algorithmic}
\usepackage{array}
\usepackage[caption=false,font=normalsize,labelfont=sf,textfont=sf]{subfig}
\usepackage{textcomp}
\usepackage{stfloats}
\usepackage{url}
\usepackage{verbatim}
\usepackage{graphicx}
\def\BibTeX{{\rm B\kern-.05em{\sc i\kern-.025em b}\kern-.08em
    T\kern-.1667em\lower.7ex\hbox{E}\kern-.125emX}}
\usepackage{balance}
\begin{document}
\title{Neuroadaptation in Physical Human-Robot Collaboration}
\author{Avinash Singh, Member, IEEE, Dikai Liu, Senior Member, IEEE, Chin-Teng Lin, Fellow, IEEE
\thanks{Avinash Singh and Chin-Teng Lin are with with the Australian Artificial Intelligence Institute, School of Computer Science, Faculty of Engineering and Information Technology, University of Technology Sydney, 81 Broadway, Ultimo NSW 2007, Australia. 
Dikai Liu is with the Centre for Autonomous Systems, Faculty of Engineering and Information Technology, University of Technology Sydney, 81 Broadway, Ultimo NSW 2007, Australia.(Corresponding email: avinash.singh@uts.edu.au)}}


\markboth{Journal of \LaTeX\ Class Files,~Vol.~xx, No.xx, September~2023}%
{How to Use the IEEEtran \LaTeX \ Templates}

\maketitle

\begin{abstract}
Robots for physical Human-Robot Collaboration (pHRC) systems need to change their behavior and how they operate in consideration of several factors, such as the performance and intention of a human co-worker and the capabilities of different human-co-workers in collision avoidance and singularity of the robot operation. As the system's admittance becomes variable throughout the workspace, a potential solution is to tune the interaction forces and control the parameters based on the operator’s requirements. To overcome this issue, we have demonstrated a novel closed-loop-neuroadaptive framework for pHRC. We have applied cognitive conflict information in a closed-loop manner, with the help of reinforcement learning, to adapt to robot strategy and compare this with open-loop settings. The experiment results show that the closed-loop-based neuroadaptive framework successfully reduces the level of cognitive conflict during pHRC, consequently increasing the smoothness/intuitiveness of human-robot collaboration. These results suggest the feasibility of a neuroadaptive approach for future pHRC control systems through electroencephalogram (EEG) signals.
\end{abstract}

\begin{IEEEkeywords}
physical human robot collaboration, reinforcement learning, cognitive conflict, deep learning, electroencephalogram
\end{IEEEkeywords}

\section{Introduction}
\IEEEPARstart{S}{hared} control or shared autonomy plays an important role in collaborative technologies. Collaborative robots or cobots or Physical Human-Robot Collaboration (pHRC)~\cite{ref1} is one of such example. The pHRC happens when a human and a robot contact and exchange forces to accomplish a common task. In such a scenario, the human is always in complete or partial control of the robot’s motions. Some of the examples of pHRC systems include a cobot for material handling~\cite{ref2}, a human-robot system for homokinetic joint assembly~\cite{ref3}, a wearable robot tested for lifting and holding tasks~\cite{ref4}, hand rehabilitation~\cite{ref5}, a lower-limb exoskeleton~\cite{ref6} and an exoskeleton to rehabilitate shoulder and elbow~\cite{ref7}. The collaborative robotics world is currently undergoing fundamental paradigm shifts in terms of research and applications~\cite{ref8}. One of the most common directions in pHRC is a human-centered design of robot mechanics and control~\cite{ref9}. However, it has limitations in terms of its understanding of the human co-worker’s needs which stem mainly from subjective opinion~\cite{ref10}. These pHRCs do not take into account that every human co-worker is unique in their skills, knowledge, and experience. Interaction dynamics that are suitable for one human co-worker might be uncomfortable for others. However, in current pHRC settings~\cite{ref11,ref12,ref13}, human usually adapt over time to any gaps between their expectations and robot behavior~\cite{ref14}. When such an adaptation has already reached a stagnant stage where further improvement is not possible, it might create a significant delay in the actual process and performance will drop until and unless the human co-worker has adapted to the robot. All existing pHRC~\cite{ref3,ref9,ref12,ref15} share the common requirement of such adaptation from human co-workers for close, safe, and dependable physical interaction in a shared workspace. However, as we reach a point where such robots need to be carefully designed for the human co-workers’ need, we now ask, What if such an adaptation is taken care of by the robot itself rather than the human co-worker? A pHRC system that can safely sense, reason, learn and act while safely working in a shared workspace with human is a sound aim. However, designing a pHRC system that can adapt to an human co-workers’ needs, change and performance require a better understanding of the human co-worker’s feelings/thoughts during their collaboration with a robot.
To design a pHRC that can couple with control, planning, and learning according to human needs, performance, and capability in real-time, understanding the changes of human co-worker’s cognitive state may provide the best source of information that can be utilized to tune and adapt the robots in pHRC settings. Recent advances in wearable physiological sensors such as electromyogram (EMG), heart-rate variability (HRV), electroencephalogram (EEG), photoplethysmogram (PPG), etc., have given rise to new paradigms in understanding a human’s cognitive state changes during human-robot interaction~\cite{ref16}.  The researchers now are using them in different settings to enhance the understanding of human co-worker’s states that include stress~\cite{ref17, ref18}, attention~\cite{ref19}, mental fatigue~\cite{ref20}, engagement~\cite{ref21}, workload~\cite{ref22} in human-robot collaboration and interaction environments. The information relating to such cognitive states becomes important if a pHRC task creates undue mental workload, stress, or mental fatigue on human co-workers. However, these cognitive states are not very suitable for understanding the human co-workers’ covert cognitive state as influenced by collaborating with a robot. Also, mental fatigue, stress, and workload develop gradually and do not necessarily reflect the ongoing changes in human co-worker’s feelings. Therefore, tracking this continuously changing cognitive state and feelings during collaborating with a robot remains highly desirable. 

\begin{figure*}[ht]
\centering
\includegraphics[width=\textwidth]{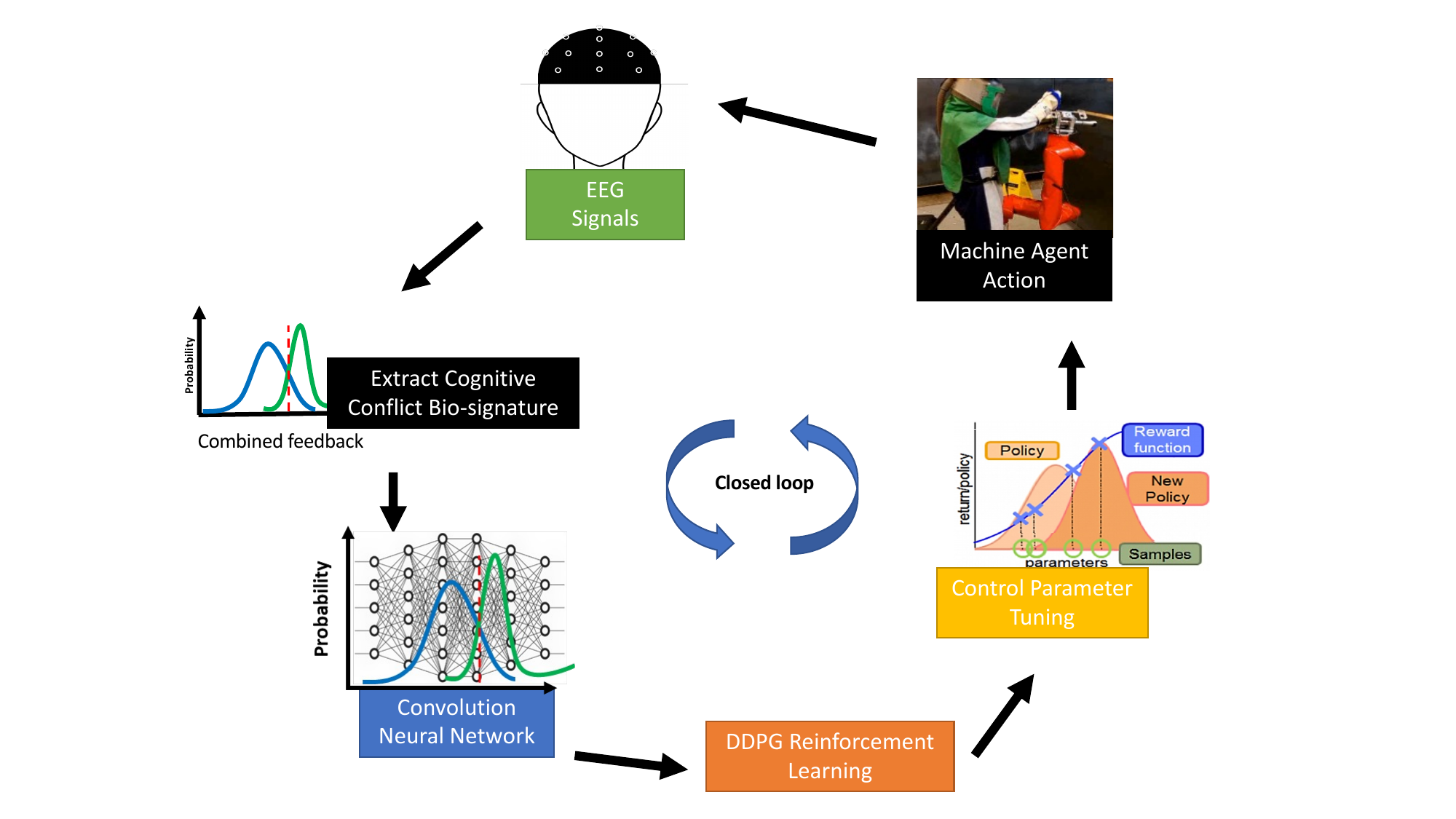}
\caption{A schematics diagram for the Neuroadaptive framework for pHRC.}
\label{fig1}
\end{figure*}

One such relevant cognitive factor here is cognitive conflict. This occurs if the predicted outcome of a human co-worker’s desirable action during collaborating with a robot does not match the robot’s eventual action. It has been utilized as one of the essential cognitive features in successfully adapting human feelings to different environmental settings~\cite{ref23, ref24}. Such an adaption of human feeling is generally known as  neuroadaptation~\cite{ref25}. Several works on neuroadaptation~\cite{ref26,ref27, ref28, ref29} have employed cognitive conflict-based features to interface with a robot to correct robot’s mistakes. However, most of these works are limited in that the human co-worker is simply observing the robot’s action from a distance i.e., non-contact action on a computer screen or using a keyboard/joystick~\cite{ref29} to control it, rather than exchanging forces and coupled motions as in pHRC settings. Furthermore, these works only focus on correcting a robot’s mistakes after a task is completed rather than truly neuroadapting to the human co-worker’s feelings during conducting the task~\cite{ref30}. In our previous work~\cite{ref31}, we demonstrated the possibility of detecting cognitive conflict in pHRC settings and were successfully able to decode the human co-worker’s feelings regarding the robot’s actions. We also established that cognitive conflict could be detected more readily in line with increasing the degrees of a robot’s movements, such as moving between one-dimensional (1D) and two-dimensional (2D) settings~\cite{ref31}. Continually working on this line of work, this paper presents a novel neuroadaptive framework for pHRC to adapt the human co-worker’s feeling while physically collaborating with a robot in real-time within a short duration of ~15 minutes, as shown in Figure~\ref{fig1}.

\section{Materials and Methods}

\subsection{Experiment design}
A pHRC robot called the ANBOT~\cite{ref45} was used as a platform to run this experiment. It consists of a UR10 (Universal Robots) controlled with an admittance-based controller via a force/torque sensor mounted between the arm and the tool. We used a 32-channels EEG device called MOVE (Brain Product GmbH) to record the brain activity, with a sampling rate of 1000Hz. The electrodes were placed on the scalp with an impedance below 50  and accordingly to the 10-20 international system~\cite{ref46}.

\begin{figure}[!t]
\centering
\includegraphics[width=\columnwidth]{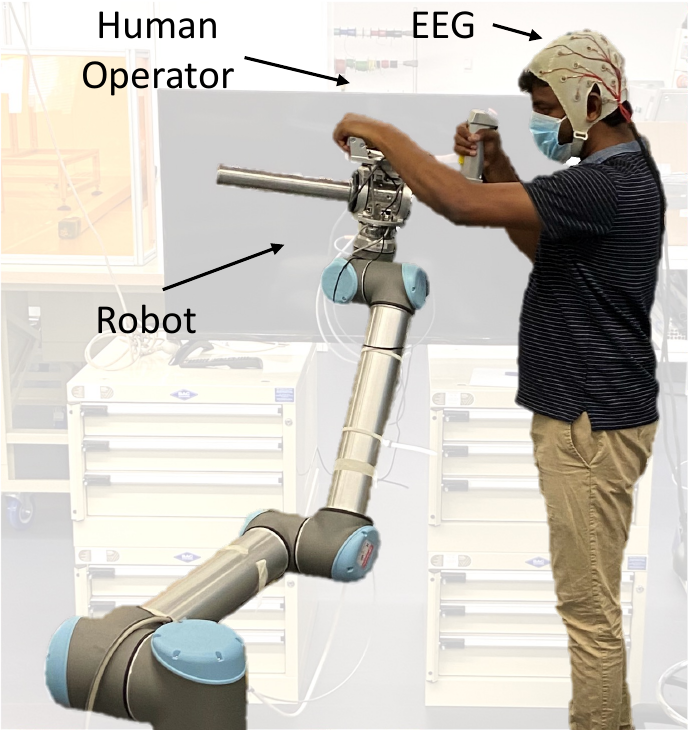}
\caption{Experiment Design. A participant is controlling pHRC in a singularity experiment wearing a wired EEG cap .}
\label{fig2}
\end{figure}

Before starting the experiment (see setup in Figure~\ref{fig2}), we recorded baseline data with the participants sitting still, first with open eyes and then with closed eyes. Each baseline lasts about one minute, and the experiment is divided into two parts: open-loop (without neuroadaptive framework) and closed-loop (with neuroadaptive framework). In the open-loop condition, participants explored the shared workspace. They were explicitly asked to stretch joint three of the UR10 (robot elbow) and test the proximity of singularity. The damping variables \(\ \bar\sigma_0\) and \(\bar\sigma_1\) are in bounded to the range [0.35, 0.45], when the robot is approaching a singular configuration and [0.25, 0.45] when it is moving away from the singularity following Carmichael et al.~\cite{ref15}.  

Similarly, in the closed-loop conditions, the participants operate the ANBOT while, concurrently, a reinforcement learning algorithm is setting and adjusting \(\bar\sigma_1\) automatically, while \(\bar\sigma_0\) is fixed and equal to 0.25. \(\bar\sigma_1\) bounded to the range [0.35, 0.45], when the robot is approaching a singular configuration and bounded to [0.25, 0.45], when moving away from the singularity similar to the open-loop condition. This ensures that there is always some damping when approaching singularity (fast to slow), while there can be no damping to slow damping if leaving singularity. 
At the end of both the conditions, we asked the participants to fill a questionnaire. In this questionnaire, we asked the participants' preference for the two conditions. Each open and closed-loop condition took about ~15 minutes, excluding the time to prepare the participant for an experiment.

\subsection{Participants}

The cohort for this experiment consisted of fourteen healthy participants – three females and eleven males, aged between 18 and 42. Two participants were, however, excluded from the EEG data analysis because the data was corrupted. The two excluded participants are, nevertheless, included in the questionnaire about the open-loop condition. Before participating in the study, each participant was given a full explanation of the experimental procedure, and each provided informed consent. Ethics approval was issued by the Human Research Ethics Committee of the University of Technology Sydney, Australia. The experiment was conducted in a large room by a male experimenter. None of the participants had a history of neurological or psychological disorders, which could have affected the experiment results. All the participants were allowed to wear glasses for corrected vision.

\subsection{Methodology}
The EEG data were processed offline to see whether a cognitive conflict existed for specific conditions. We used a Lab-Streaming Layer (LSL)\footnote{https://github.com/sccn/labstreaminglayer} to synchronize the data with event markers defining the conditions. The event markers were sent for different values of \(\sigma_i\), particularly for \(\sigma_i < \sigma_q\), \(\sigma_i < (\sigma_1-\sigma_0)/2\), \(\sigma_i < (\sigma_1-\sigma_0)/4\) and \(\sigma_i < (\sigma_1-\sigma_0)/8\). The EEG data were first resampled to 250Hz and filtered with a band-pass filter in the range of 2-50Hz. The process then employed Artifact Subspace Reconstruction (ASR)~\cite{ref47} to automatically reject bad channels, followed by independent component analysis (ICA)~\cite{ref32}. We next applied ADJUST~\cite{ref48} to further clean the data from artifacts. The epochs were then extracted to compute the Power Spectral Density (PSD) (49). Epochs are extracted for each condition from the relative event marker to 400ms after that. The resultant epoched data have been divided into three segments: the first 1/3, second 1/3 and third 1/3. The PSD of the recorded baseline for open eyes is also computed after this data went through the same pipeline. The baseline PSD is then subtracted from the PSD of each condition for participants. The result is a normalized PSD for each participant to allow for considerable variation in PSD between three time-segments. 

\subsection{Cognitive Conflict Deep Deterministic Policy Gradient (CC-DDPG) model}

\begin{figure*}[ht]
\centering
\includegraphics[width=\textwidth]{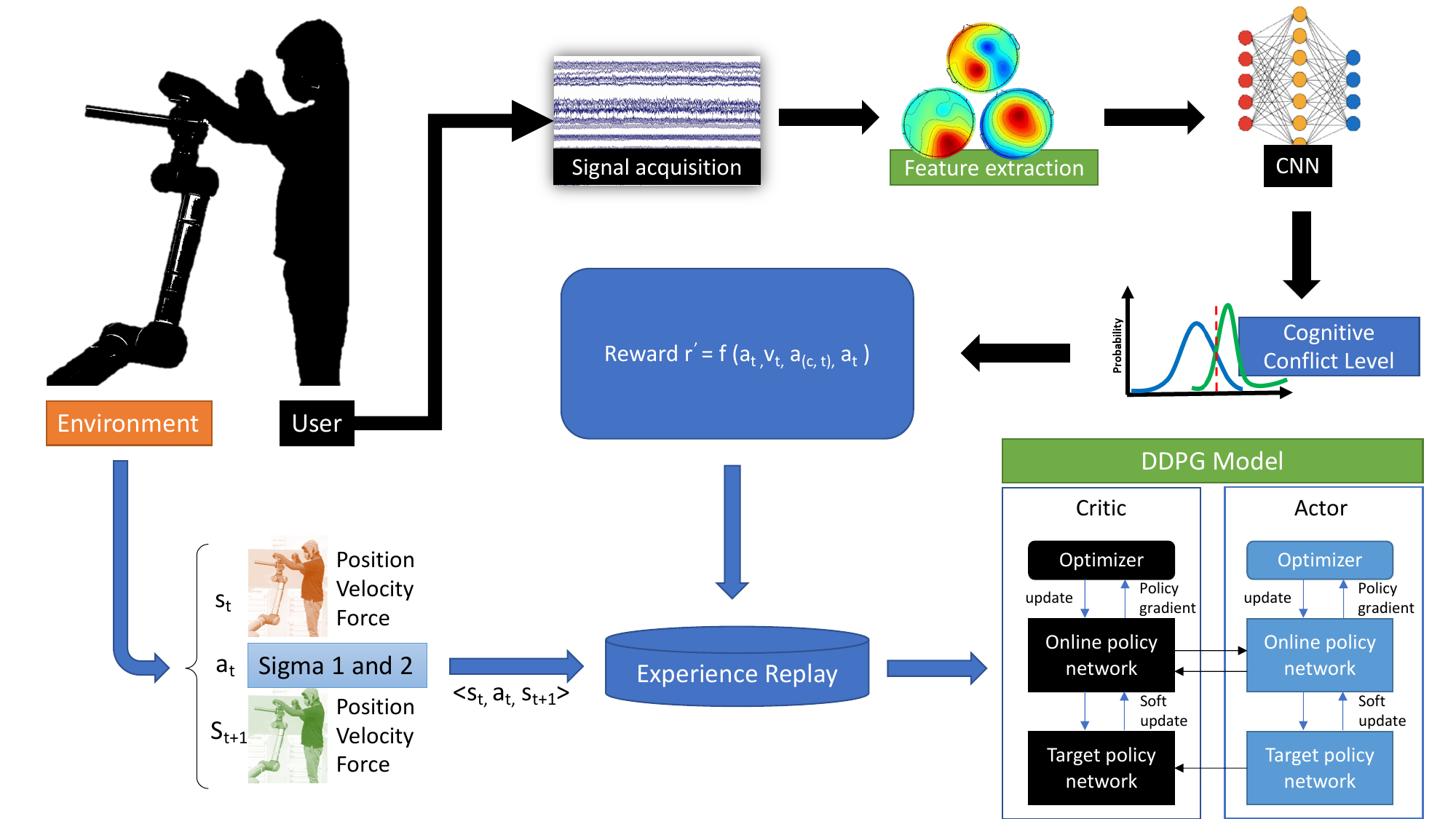}
\caption{Schematic model structure for neuroadaptive framework using CC-DDPG.}
\label{fig3}
\end{figure*}

Motivated by the human goal-directed learning process, reinforcement learning models are based on the interaction between environment and agent by balancing the exploration and exploitation by maximizing the return (i.e., rewards). In this paper, a neuroadaptive framework for pHRC was develop based on deep reinforcement learning (DRL)~\cite{ref50} (see Figure~\ref{fig3}). DRL can generate control policies directed from the robot’s states for pHRC. In pHRC, the robot’s state consists of the magnitude of position, velocity, and force applied by the participant while collaborating. Theoretically, the state-space for each variable is unlimited in this case. To solve such DRL problems requires a DRL for continuous action space, leading to the choice of a deep deterministic policy gradient (DDPG)~\cite{ref43} algorithm to learn environments with a continuous flow of state and action. 

Let’s assume the actor is \(\pi_\mu\ (a_t | s_t)\) and the \(critic\ Q(s_t, a_t |\theta)\). The actor takes \(s_t\) and produces the action \(a_t\), while the critic takes these actions \(a_t\) and \(s_t\) and produces \(S(s_t,a_t)\)\ to minimize the reward-prediction error (RPE) as follows:  

\begin{equation} 
\centering
\label{equ1}
\delta_t=r_t+\gamma Q(s_(t+1),\pi(s_(t+1))|\theta)-Q(s_t,a_t\ |\theta)
\end{equation}

where loss function theta is as follows:	

\begin{equation} 
\centering
\label{equ2}
L=1/N\ \sum_i(y_i-Q(s_i,a_i\ |\theta))^2 
\end{equation}

While the actor following the policy gradient theorem as follows:

\begin{equation} 
\centering
\label{equ3}
\nabla_\mu\ \pi(s,a)=E_\rho(s)\ \ [\nabla_a\ Q(s,a | \theta)\ \nabla_\mu\ \pi(s|\mu)]
\end{equation}

\begin{equation} 
\centering
\begin{split}
\label{equ4}
\nabla_w\ \pi(s_i\ )\approx 1/N\ \sum_i \\
\nabla_a\ Q(s,a | \theta)\ |_(s=s_i,a=\pi(s_i ) ) \nabla_w pi(s)|_(s=s_i )
\end{split}
\end{equation}

\subsection{Reward function with Cognitive Conflict}

The reward setting of the reinforcement learning (RL) algorithm is a highly essential component for effective model training. Improper reward setting relates directly to a failure of the model. Russell et al.~\cite{ref53} give an example: when the reward of a vacuum cleaner is set to “absorb dust”, vacuum cleaners will get rewards by “spraying dust” and then “absorbing dust”. The sparse reward may lead to inefficient model exploration and learning, slow iteration, and even difficulty in converging. We need to model the participant’s feeling of cognitive conflict as measured directly from the brain as a reward to optimize the robot’s action. We have formulated the reward function as follows:

\begin{equation} 
\centering
\label{equ5}
r=r^\prime+r_{CNN}
\end{equation}

where r represents the reward from cognitive conflict for each action performed between a participant and a pHRC. \(r_{CNN}\) is used to describe the comprehensive evaluation of a participant’s feeling derived from 32-channels EEG over the 1.2s period. 

The \(r_{CNN}\) used a convolution neural network (CNN) based on DeepConvNet~\cite{ref54} to classify three levels of cognitive conflict. The model is trained on three classes on previously collected cognitive conflict data in~\cite{ref55}, where three classes were normal conditions such as no conflict condition, slow conflict condition, and sudden conflict condition. We have assigned \(r_{CNN}\) as 100, 50, -100 for normal conflict, slow conflict, and sudden conflict respectively after each classification over the 1.2s second of a trial.  The data used for DeepConvNet involved minimal pre-processing. The acquired EEG data were filtered with a band-pass filter in the range of 2 to 50Hz before extracting the epoch of 1.2s long in real-time by buffering data of 1200 samples using LSL. Similarly, the robot data used for CC-DDPG consisted of the human-robot collaborating forces, position, end-effector velocities, and sigma values. This data was sampled by 125 Hz. See more detail about the DeepConvNet model, its hyperparameters, and training data results in Supplementary results.

\section{Results}

\begin{figure*}[!t]
\centering
\includegraphics[width=\textwidth]{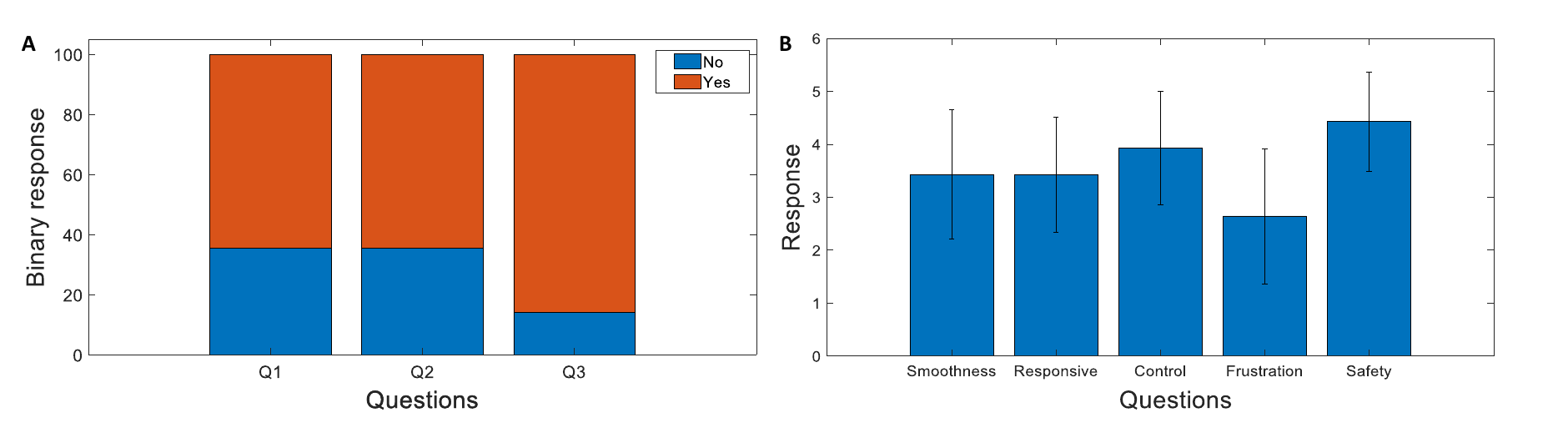}
\caption{The questionnaire results from participants comparing closed-loop and open-loop neuroadaptive frameworks. (A) The results show the participants’ experience toward the pHRC on a binary scale on three questions: “Q1. Have you ever used a robot arm before?”; “Q2: Do you know how robot arms work?”, and “Q3: Do you know what robot kinematic singularity is ?”; (B) The results show the participants’ experience toward pHRC’s smoothness, responsiveness, control, frustration, and safety on the five-point Likert scale (with standard deviation).}
\label{fig4}
\end{figure*}

We have evaluated the neuroadaptive framework for pHRC in fourteen participants performing a singularity task~\cite{ref31} with a neuroadaptive framework (closed-loop) and without (open-loop) conditions. This task is to freely move and stretch the three joints of the UR10  (robot’s elbow)\footnote{  https://www.universal-robots.com/products/ur10-robot/} and test the proximity of singularity. These participants had minimal experience with robot control, as reported in the questionnaire (see Figure~\ref{fig4}).

\subsection{Comparison in behavioral results for open and closed-loop-neuroadaptive frameworks}

Figure~\ref{fig4}(A) shows that about 86\% of participants did not know the type of task they were going to perform before the experiments, 64\% had never used any robotic arms before and did not know how a robotic arm should work. Figure~\ref{fig4}(B) indicates how participants rated the importance of their feelings on the questions of smoothness – 69\%, responsiveness – 69\%, control – 79\%, frustration – 53\%, and safety 89\%, while using a closed-loop neuroadaptive framework condition compared to an open-loop condition.

\begin{figure*}[!t]
\centering
\includegraphics[width=\textwidth]{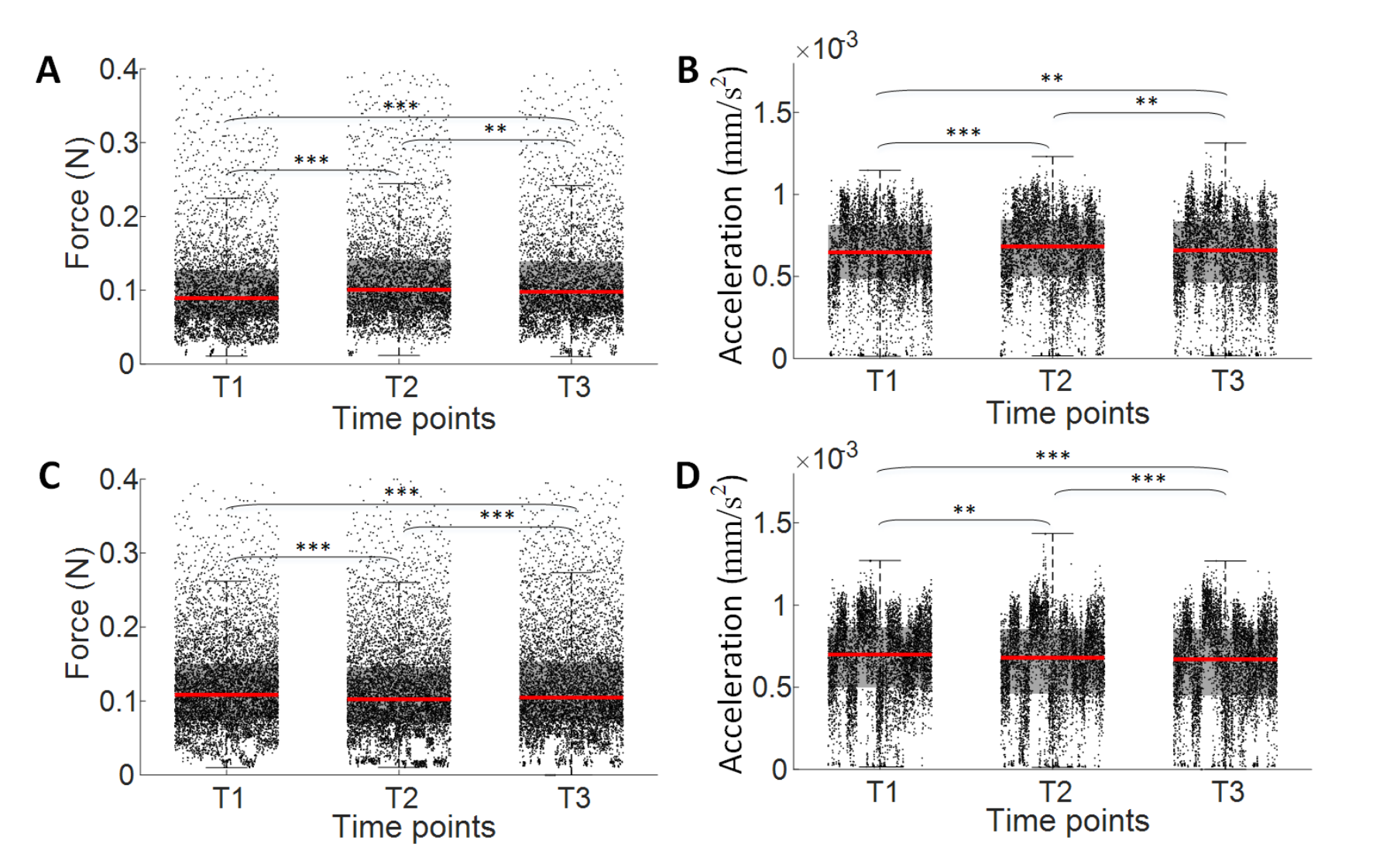}
\caption{Force and acceleration while participants control it in open and closed-loop condition (A) The force applied by the participants over the three time-segments (T1, T2, and T3) over the open-loop condition; (B) The acceleration of robotic arm over the three time-segments in open-loop condition; (C) The force applied by the participant over the three time-segments over the closed-loop condition; (B) The acceleration of the robotic arm over the three time-segments in a closed-loop condition. (Bar plots indicate mean ± SEM. Statistical analysis using a two-way repeated-measures ANOVA. *p< 0.05, **p<0.005, ***p<0.0005.).}
\label{fig5}
\end{figure*}

We divided the data into three equal time segments as T1 (first 1/3), T2 (second 1/3), and T3 (third 1/3). The divided data then extracted and processed the forces applied by participants during pHRC, and the acceleration of the robotic arm in open and closed-loop conditions as shown in Figure~\ref{fig5}. A repeated-measure ANOVA was conducted to compare 1 (force data) x 3 (time-segments) and 1 (acceleration data) x 3 (time-segments) for open and closed-loop conditions. There was a significant difference between the force at three time-segments (F (2, 9154) = 77.938, p = .000) for the open-loop condition and closed-loop conditions (F (2, 14562) = 27.776, p = .000).  LSD post-hoc test on time-segments for open-loop revealed that there was a significant difference from T1 to T2 (p = .000), T1 to T3 (p = .000), and T2 to T3 (p =.001). Similarly, the LSD post-hoc test on time-segment under closed-loop conditions revealed also that there was a significant difference from T1 to T2 (p = .000), T1 to T3 (p = .000), and T2 to T3 (p =.000).

A repeated-measure ANOVA was also conducted to compare 1 (acceleration data) x 3 (time-segments) for open and closed-loop conditions. There was a significant difference between the force at three time-segments (F (2, 8096) = 15.768, p = .000) for an open-loop condition as well as (F (2, 12108) = 20.301, p = .000) for closed-loop conditions.  LSD post-hoc test on time-segments for open-loop condition revealed that there was significant difference from T1 to T2 (p = .000), T1 to T3 (p = .007), and T2 to T3 (p = .004). Similarly, LSD post-hoc testing on time-segment for closed-loop terms also revealed that there was significant difference from T1 to T2 (p = .008), T1 to T3 (p = .000), and T2 to T3 (p = .000).

\subsection{The comparison between EEG results for open and closed-loop-neuroadaptive framework}

\begin{figure*}[ht]
\centering
\includegraphics[width=\textwidth]{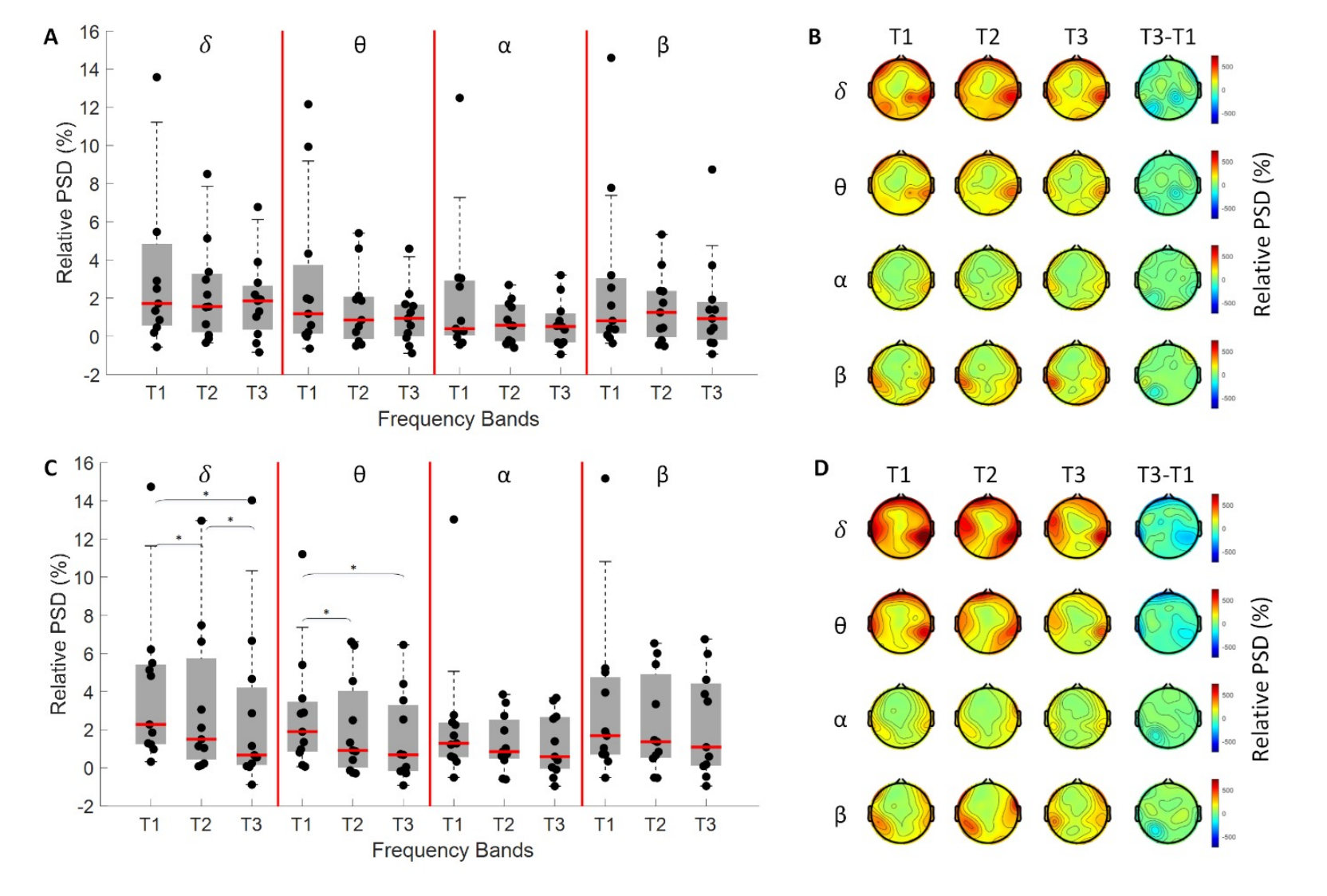}
\caption{Power Spectral Density at delta, theta, alpha, and beta for the open and closed-loop condition within the participant at three time-segements. (A) Boxplot in open-loop condition at times T1, T2, and T3 at channel ‘Fz’; (B). Topoplots open-loop condition at times T1, T2, and T3; (C) Topoplots with the closed-loop condition at times T1, T2, and T3; (D) Boxplot in closed-loop condition at times T1, T2, and T3 at channel ‘Fz’. (Bar plots indicate mean ± SEM. Statistical analysis using a two-way repeated-measures ANOVA. *p< 0.05, **p<0.005, ***p<0.0005.)}
\label{fig6}
\end{figure*}

To understand if the participants’ brain dynamics exhibit any difference(s) when collaborating with pHRC both with (closed-loop) and without (open-loop) neuroadaptive framework, we also divided the data into three-time segments (T1, T2, and T3) and extracted cognitive conflict-related neuro-markers based on power-spectral density (PSD) for both conditions as presented in Figures~\ref{fig6}~(A) and~\ref{fig6}~(C). A repeated-measure ANOVA was conducted to compare 4 (power bands) x 3 (time-segments) for closed-loop settings. There was no significant difference for PSD between power bands (F (3, 30) = 0.747, p = .532), but there was a significant difference in PSD between time-segments (F (2, 20) = 4.431, p = .026).  LSD post-hoc test on time-segments (T1, T2, and T3) for all bands revealed that there was significant difference in PSD from T1 to T2 (p = .043), T1 to T3 (p = .030), and T2 to T3 (p = .039) for delta bands, and T1 to T2 (p = .050), T1 to T3 (p = .029), and T2 to T3 (p = .055) for theta bands. However, there was no significant difference in PSD from T1 to T2 (p = .130), T1 to T3 (p = .122), and T2 to T3 (p = .682) for alpha bands, and T1 to T2 (p = .109), T1 to T3 (p = .117), and T2 to T3 (p = .978) for beta bands. 

A repeated-measure ANOVA was also conducted to compare PSD for 4 (power bands) x 3 (time-segments) for open-loop settings. There was a no significant difference in PSD between power bands (F (3, 30) = 0.247, p = .863), but there was a significant difference in PSD between time-segments (F (2, 20) = 4.050, p = .033).  LSD post-hoc testing on time-segment (T1, T2, and T3) for all bands revealed that there was no significant difference in PSD from T1 to T2 (p = .073), T1 to T3 (p = .065), or T2 to T3 (p = .350) for delta bands; T1 to T2 (p = .081), T1 to T3 (p = .066), and T2 to T3 (p = .243) for theta bands, T1 to T2 (p = .156), T1 to T3 (p = .149), and T2 to T3 (p = .643) for alpha bands, and T1 to T2 (p = .136), T1 to T3 (p = .140), and T2 to T3 (p = .882) for beta bands. 

In addition to looking at only one EEG channel (Fz) in the frontal lobe following previous work~\cite{ref24}, we observed PSD topoplots utilizing all the EEG channels as shown in Figures~\ref{fig6} (B) and ~\ref{fig6} (D). We also compared topoplots from the participant(s) at three time-segments (T1, T2, and T3) for two conditions. A repeated-measure ANOVA was also conducted to compare PSD for topoplots in 4 (power bands) x 3 (time-segments) for closed-loop settings. As the data in Figure~\ref{fig6}(D) indicate, there was a significant difference between power bands (F (3, 90) = 67.995, p =.000) as well as time-segments (F (2, 60) = 24.657, p = .000) for topoplots. An LSD post-hoc test on time-segments (T1, T2, and T3) for all bands revealed that there was significant difference in PSD for topoplots from T1 to T2 (p = .032), T1 to T3 (p = .000), and T2 to T3 (p = .000) for delta bands; T1 to T2 (p = .031), T1 to T3 (p = .000), and T2 to T3 (p =.000) for theta bands; T1 to T3 (p = .003), and T2 to T3 (p = .003) for alpha bands, and T1 to T2 (p = .002), T2 to T3 (p = .017) for beta bands. However, there was no significant difference in PSD for topoplots from T1 to T2 (p = .338) in alpha bands or T1 to T3 (p = .459) in beta bands. 

Similarly, a repeated-measure ANOVA was conducted to compare 4 (power bands) x 3 (time-segments) in open-loop settings in PSD for topoplots. As illustrated in Figure~\ref{fig6}(B) (topoplots), there was a significant difference between power bands (F (3, 90) = 75.328, p = .000) as well as time-segments (F (2, 60) = 6.501, p = .003) in PSD for topoplots. LSD post-hoc testing on time-segments (T1, T2, and T3) for all bands revealed that there was significant difference in PSD for topoplots from T1 to T3 (p = .000), and T2 to T3 (p = .000) in delta bands, T1 to T3 (p = .000), and T2 to T3 (p = .000) in theta bands. However, there was no significant difference in PSD for topoplots from T1 to T2 (p = .354) in delta bands; T1 to T2 (p = .074) in theta bands; T1 to T2 (p = .088), and T1 to T3 (p = .547), T2 to T2 (p = .455) in alpha bands, and T1 to T2 (p = .629), T1 to T3 (p = .115), T2 to T3 (p = .114) for beta bands. 

\begin{figure}[!t]
\centering
\includegraphics[width=\columnwidth]{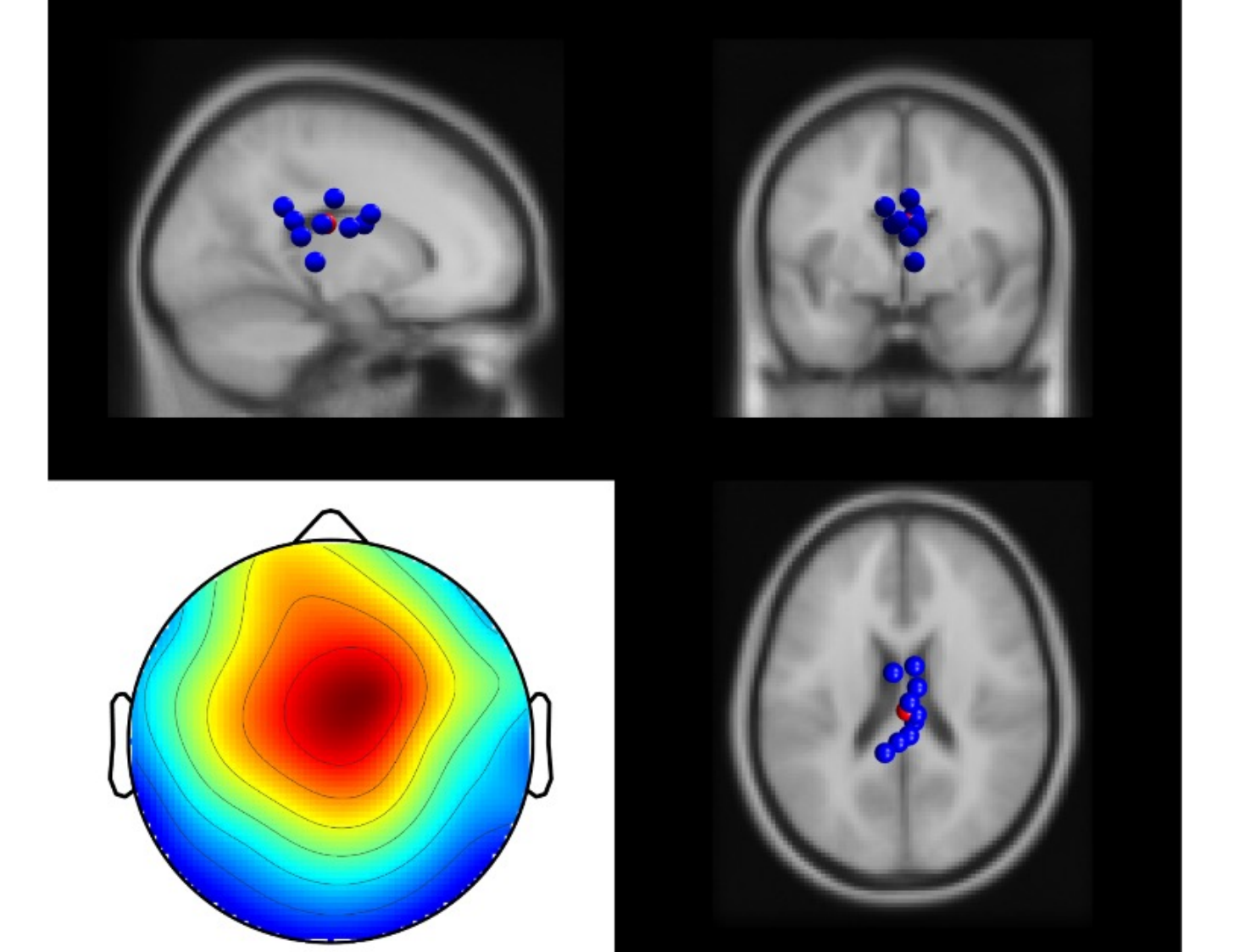}
\caption{Origination of cognitive conflict. Anterior Cingulate Cortex (ACC) and relative dipole positions in three views.)}
\label{fig7}
\end{figure}

To further evaluate the brain dynamics, we have utilized the Independent Component Analysis (ICA)~\cite{ref32} with dipole-fitting~\cite{ref33}  to determine if cognitive conflict originates from the anterior cingulate cortex~\cite{ref34} . We found this is indeed the case, as shown in Figure~\ref{fig7}. It clearly indicates that cognitive conflict originates in the ACC region of the brain, as dipole-fitting analysis and Tailarch coordinates verify. It is important to note that this ACC is the same for both open and closed-loop conditions.

\subsection{Reinforcement Learning results from the closed-loop-neuroadaptive framework}

\begin{figure*}[!t]
\centering
\includegraphics[width=\textwidth]{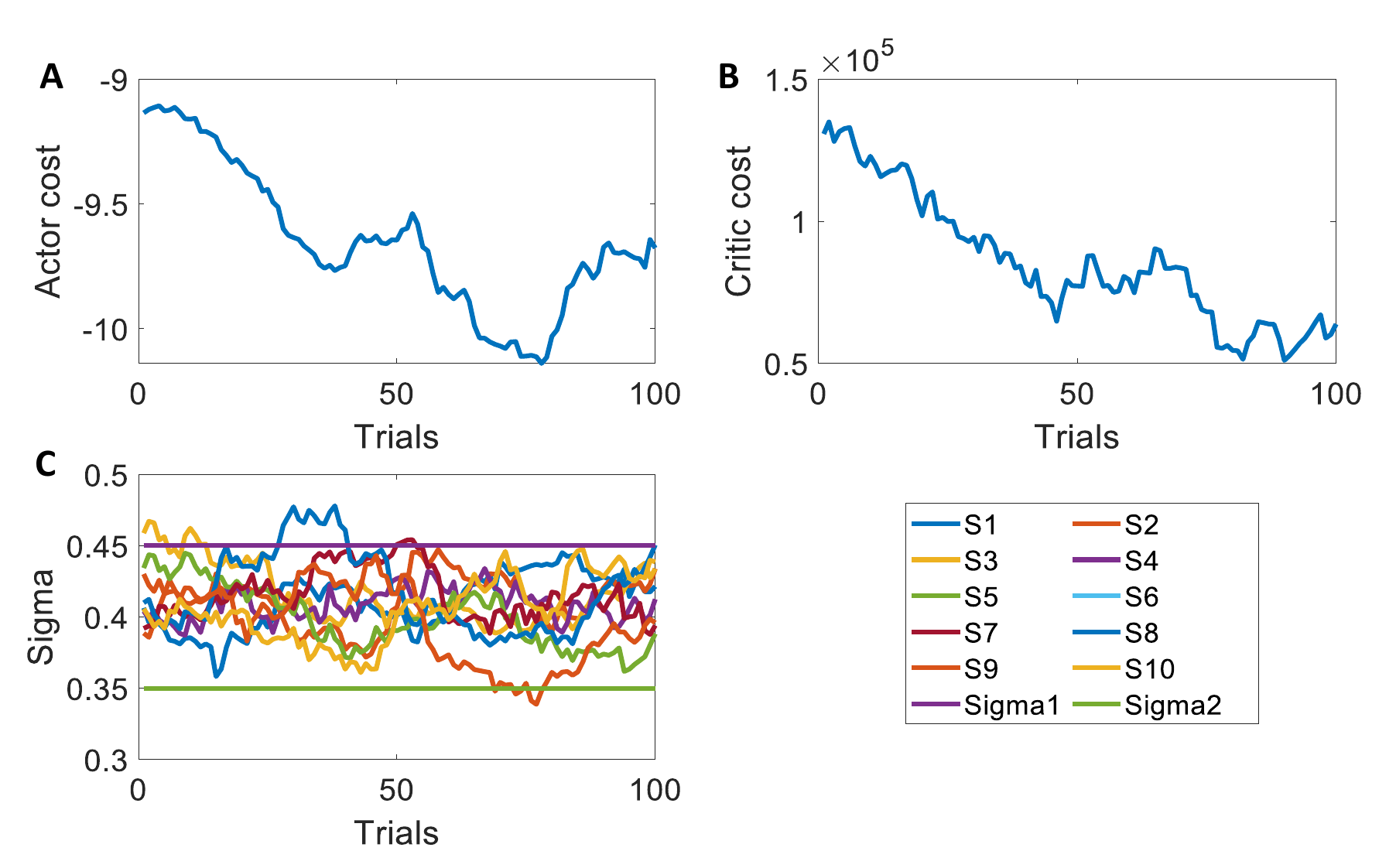}
\caption{The results from the cognitive conflict-based CC-DDPG algorithm. (A) Actor cost over 100 trials or episodes of reinforcement learning; (B) Critic cost over 100 trials or episodes of reinforcement learning; (C) Action was taken by CC-DPPG for all participants separately over the 100 trials or episodes of reinforcement learning.}
\label{fig8}
\end{figure*}

To reveal the efficacy of the cognitive-conflict deep deterministic policy gradient (CC-DDPG) algorithm, we evaluated a representative participant's actor and critic costs. We also assess the action taken by the CC-DDPG algorithm for all participants to neuroadaptive pHRC. As Figure~\ref{fig8} (A) illustrates, throughout 100 trials (i.e., games), the actor cost sharply decreases as we approach 40 trials and then shows a slight increase beyond this up to 50 trials, before being followed by a sharp decrease up to 90 trials, and a slight increase

in the last 20 trials. Similarly, for critic cost as plotted in Figure~\ref{fig8} (B), the cost decreases as we rise to around 50 trials, followed by a slight increase before a decrease at 70 trials. Then, a sharp decrease of up to 80 games ensued by a slight increase that is succeeded by a decline in the next 20 trials. Overall, both actor and critic costs decrease over the trials, but initial convergence is achieved at around 40 trials.

In normal conditions, without reinforcement learning (RL), the standard sigma values are set between 0.35 (\(\sigma_1\)) to 0.45 (\(\sigma_2\)) and CC-DDPG is required to decide a value that minimizes cognitive conflict. As we show in Figure~\ref{fig8}(C), participant preference generally moves more toward the higher sigma values, which also vary for each participant at different instances of the games. In comparison with actor and critic costs around trials 40 and 100, sigma values approach Sigma2.

\section{Discussion}
The present work demonstrates a novel closed-loop-neuroadaptive framework for pHRC. The results show that cognitive conflict is an effective neuro-marker that enables the information about human co-workers’ feelings while collaborating with the robot used in neuroadaptation. The system as developed through the work of this paper successfully demonstrates its effectiveness as seen by PSD changes in brain dynamics and topoplots. The results further strengthen the efficacy of the neuroadaptive framework through human co-workers’ behavior reflected by force and acceleration data in addition to participant’s feedback as recorded by robot’s sensors, questionnaires, and the output from the CC-DDPG algorithm.  

In work presented here, most participants had not been exposed to this or similar mechanisms before. Nevertheless, they were familiar with robot interaction but not of a pHRC kind. Therefore, they show almost equal attitudes towards both open and closed loops as recorded by questionnaires in terms of smooth and responsive control in pHRC settings. However, participants prefer safety and control in a closed-loop-neuroadaptive framework over an open-loop condition, potentially due to continuous adaptation compared to an open-loop condition. Another potential reason for better safety and control in closed-loop conditions, compared to the open-loop counterpart, might be a participant's perception of how a robot reacts to a too sudden change in sigma values, these require a participant’s continuous change in strategy, hence encouraging them to feel safer and in control. This is also in alignment with previous studies~\cite{ref35} of pHRC systems where novice participants felt less resistive toward control, thus more control due to continuous changes in impedance level of robot.

It was interesting to note how participants also felt highly frustrated for closed-loop settings, which seems obvious again because they were continuously prompted with change to learn about their feelings by the CC-DDPG algorithm. A similar trend is visible in the force and acceleration applied by participants to collaborate with a robotic arm. It seems clear that, during the adaptation process, participants were able to control the robot much more smoothly and that flexible hand movement allows more stable acceleration. It can be argued that such a reduction in the force applied over time and a change in acceleration is due to the effect of the participant learning. If that is the case, the open-loop condition should have delivered similar findings, something that was not depicted in the results. There is a learning process involved, although not human, but robotic, based on human’s feelings with the help of a neuroadaptive framework.  

Looking closely at the human brain with the help of EEG, it became even more apparent that our hypothesis on neuroadaptation was correct. The result from the PSD depicts a reduction in power over time (T1, T2, and T3) for delta and theta bands. The delta power band is known mainly for modulation while monitoring and adapting to external stimuli, as shown by~\cite{ref36} . The reduction in the theta band notably originated in the ACC areas of the brain~\cite{ref37}, known to be related to a reduction in observation error in the environment. Following the theory of cognitive conflict~\cite{ref38} , our results suggest that, as the robot shapes the participants’ roles, the theta power originated in ACC areas is reducing. It is further evident with open-loop results on the theta band, where similar findings are not observed because there was no neuroadaptation procedure in place. The decrease in delta and theta bands is also observable in topoplots, particularly focused around frontal-central areas related which reflect ACC~\cite{ref34}  region of the brain, which aligns with previous findings in cognitive conflict~\cite{ref23, ref39}.

Another interesting point is that alpha and beta power bands demonstrate a similar effect in both open and closed-loop conditions. The reason could be that they are not generally known to be related to a participant’s cognitive states, such as cognitive conflict. Alpha power relates to changes in the level of attention~\cite{ref40}  or stimuli novelty ~\cite{ref41}, while beta power is known for change in cognitive processing~\cite{ref42}. These factors do not apply directly to pHRC settings, particularly for a performed experiment. Consequently, no changes were observed.

Participants' closed-loop neuroadaptive frameworks were further strengthened via CC-DDPG results reflected by actor and critic costs in training the model and tuning the robot’s sigma values. Over the total duration of trials, as adaptation improved, the cost for actors and critics lessened. As per the DDPG algorithm~\cite{ref43}, it is a sign that the reinforcement learning (RL) model is learning about the environment and adapting to its condition to maximize the reward represented in the form of a reduction in cognitive conflict.    

Although this work shows how neuroadaptation can be achieved using cognitive conflict in a pHRC setting, it still suffers from several limitations. 

\begin{itemize}
    \item The current setup used a 32-channel wet sensors-based MOVE system. Due to its long setup time and the common problem of gel drying out, the system is suitable only for the lab environment. We believe it should be possible to reproduce the results using off-the-shelf portable, wireless, dry-EEG devices such as Mindo~\cite{ref44}.
    \item Synchronization of the whole system is also a challenge for hardware integration, especially if cognitive conflict is a target. The classification algorithm, information streamed by a pHRC system introduces, and command issue to robot, all together create a delay in closed-loop condition. There is also a potential delay in processing all this information and sending it over an LSL system. We estimated the latency to summate to approximately 200-300ms, which might cause a delay in EEG signals and, subsequently, with cognitive conflict information. For future work that focuses on the neuroadaptation framework, dedicated synchronization hardware should be used.
    \item The participants who took part in the experiment were of limited number and age (18-42 years) and do not represent a general population sample that operates pHRC. For future work, a broader age population will be recruited for such experiments to make sure that age, gender, and experience do not influence the results.
    \item Finally, the tasks involved in this work can induce boredom in some participants. However, clearly defined tasks that require more complex physical human-robot interaction might not result in boredom. For example, in~\cite{ref24}, the participant is required to complete a task that also produces a score, thereby gamifying the interaction. We believe that the gamified setup can also be applied to the system we propose.
\end{itemize}


\bibliographystyle{ieeetr}
\bibliography{bibliography}

\begin{thebibliography}{10}

\bibitem{ref1}
J.~E. Colgate, W.~Wannasuphoprasit, and M.~A. Peshkin, ``Cobots: Robots for collaboration with human operators,'' in {\em ASME 1996 International Mechanical Engineering Congress and Exposition}, pp.~433--439, American Society of Mechanical Engineers Digital Collection, 1996.

\bibitem{ref2}
E.~Gambao, M.~Hernando, and D.~Surdilovic, ``A new generation of collaborative robots for material handling,'' in {\em ISARC. Proceedings of the International Symposium on Automation and Robotics in Construction}, vol.~29, p.~1, IAARC Publications, 2012.

\bibitem{ref3}
A.~Cherubini, R.~Passama, A.~Crosnier, A.~Lasnier, and P.~Fraisse, ``Collaborative manufacturing with physical human--robot interaction,'' {\em Robotics and Computer-Integrated Manufacturing}, vol.~40, pp.~1--13, 2016.

\bibitem{ref4}
T.~Aida, H.~Nozaki, and H.~Kobayashi, ``Development of muscle suit and application to factory laborers,'' in {\em 2009 International Conference on Mechatronics and Automation}, pp.~1027--1032, IEEE, 2009.

\bibitem{ref5}
L.~Masia, H.~Krebs, P.~Cappa, and N.~Hogan, ``Whole-arm rehabilitation following stroke: Hand module,'' in {\em The First IEEE/RAS-EMBS International Conference on Biomedical Robotics and Biomechatronics, 2006. BioRob 2006.}, pp.~1085--1089, IEEE, 2006.

\bibitem{ref6}
H.~Yu, S.~Huang, G.~Chen, Y.~Pan, and Z.~Guo, ``Human--robot interaction control of rehabilitation robots with series elastic actuators,'' {\em IEEE Transactions on Robotics}, vol.~31, no.~5, pp.~1089--1100, 2015.

\bibitem{ref7}
M.~G. Carmichael and D.~K. Liu, ``Human biomechanical model based optimal design of assistive shoulder exoskeleton,'' in {\em Field and Service Robotics: Results of the 9th International Conference}, pp.~245--258, Springer, 2015.

\bibitem{ref8}
A.~Paulikov{\'a}, Z.~Gyur{\'a}k~Babel'ov{\'a}, and M.~Ub{\'a}rov{\'a}, ``Analysis of the impact of human--cobot collaborative manufacturing implementation on the occupational health and safety and the quality requirements,'' {\em International Journal of Environmental Research and Public Health}, vol.~18, no.~4, p.~1927, 2021.

\bibitem{ref9}
F.~Rezazadegan, J.~Geng, M.~Ghirardi, G.~Menga, S.~Mure, G.~Camuncoli, and M.~Demichela, ``Risked-based design for the physical human-robot interaction (phri): An overview,'' {\em Chemical Engineering Transactions}, vol.~43, pp.~1249--1254, 2015.

\bibitem{ref10}
F.~Ficuciello, L.~Villani, and B.~Siciliano, ``Variable impedance control of redundant manipulators for intuitive human--robot physical interaction,'' {\em IEEE Transactions on Robotics}, vol.~31, no.~4, pp.~850--863, 2015.

\bibitem{ref11}
D.~Oetomo and M.~H. Ang~Jr, ``Singularity robust algorithm in serial manipulators,'' {\em Robotics and Computer-Integrated Manufacturing}, vol.~25, no.~1, pp.~122--134, 2009.

\bibitem{ref12}
B.~Navarro, A.~Fonte, P.~Fraisse, G.~Poisson, and A.~Cherubini, ``In pursuit of safety: An open-source library for physical human-robot interaction,'' {\em IEEE Robotics \& Automation Magazine}, vol.~25, no.~2, pp.~39--50, 2018.

\bibitem{ref13}
M.~Gombolay, A.~Bair, C.~Huang, and J.~Shah, ``Computational design of mixed-initiative human--robot teaming that considers human factors: situational awareness, workload, and workflow preferences,'' {\em The International journal of robotics research}, vol.~36, no.~5-7, pp.~597--617, 2017.

\bibitem{ref14}
M.~Gombolay, A.~Bair, C.~Huang, and J.~Shah, ``Computational design of mixed-initiative human--robot teaming that considers human factors: situational awareness, workload, and workflow preferences,'' {\em The International journal of robotics research}, vol.~36, no.~5-7, pp.~597--617, 2017.

\bibitem{ref15}
M.~G. Carmichael, D.~Liu, and K.~J. Waldron, ``A framework for singularity-robust manipulator control during physical human-robot interaction,'' {\em The International Journal of Robotics Research}, vol.~36, no.~5-7, pp.~861--876, 2017.

\bibitem{ref16}
M.~Alimardani and K.~Hiraki, ``Passive brain-computer interfaces for enhanced human-robot interaction,'' {\em Frontiers in Robotics and AI}, vol.~7, p.~125, 2020.

\bibitem{ref17}
P.~Rani, J.~Sims, R.~Brackin, and N.~Sarkar, ``Online stress detection using psychophysiological signals for implicit human-robot cooperation,'' {\em Robotica}, vol.~20, no.~6, pp.~673--685, 2002.

\bibitem{ref18}
H.~F. Jelinek, K.~G. August, M.~H. Imam, A.~H. Khandoker, A.~Koenig, and R.~Riener, ``Cortical response to psycho-physiological changes in auto-adaptive robot assisted gait training,'' in {\em 2011 Annual International Conference of the IEEE Engineering in Medicine and Biology Society}, pp.~7409--7412, IEEE, 2011.

\bibitem{ref19}
K.~Dufour, J.~Ocampo-Jimenez, and W.~Suleiman, ``Visual--spatial attention as a comfort measure in human--robot collaborative tasks,'' {\em Robotics and Autonomous Systems}, vol.~133, p.~103626, 2020.

\bibitem{ref20}
S.~K. Hopko, R.~Khurana, R.~K. Mehta, and P.~R. Pagilla, ``Effect of cognitive fatigue, operator sex, and robot assistance on task performance metrics, workload, and situation awareness in human-robot collaboration,'' {\em IEEE Robotics and Automation Letters}, vol.~6, no.~2, pp.~3049--3056, 2021.

\bibitem{ref21}
K.~Kompatsiari, F.~Ciardo, D.~De~Tommaso, and A.~Wykowska, ``Measuring engagement elicited by eye contact in human-robot interaction,'' in {\em 2019 IEEE/RSJ International Conference on Intelligent Robots and Systems (IROS)}, pp.~6979--6985, IEEE, 2019.

\bibitem{ref22}
A.~H. Memar and E.~T. Esfahani, ``Objective assessment of human workload in physical human-robot cooperation using brain monitoring,'' {\em ACM Transactions on Human-Robot Interaction (THRI)}, vol.~9, no.~2, pp.~1--21, 2019.

\bibitem{ref23}
A.~K. Singh, K.~Gramann, H.-T. Chen, and C.-T. Lin, ``The impact of hand movement velocity on cognitive conflict processing in a 3d object selection task in virtual reality,'' {\em NeuroImage}, vol.~226, p.~117578, 2021.

\bibitem{ref24}
S.~Aldini, A.~Akella, A.~K. Singh, Y.-K. Wang, M.~Carmichael, D.~Liu, and C.-T. Lin, ``Effect of mechanical resistance on cognitive conflict in physical human-robot collaboration,'' in {\em 2019 international conference on robotics and automation (ICRA)}, pp.~6137--6143, IEEE, 2019.

\bibitem{ref25}
T.~O. Zander, L.~R. Krol, N.~P. Birbaumer, and K.~Gramann, ``Neuroadaptive technology enables implicit cursor control based on medial prefrontal cortex activity,'' {\em Proceedings of the National Academy of Sciences}, vol.~113, no.~52, pp.~14898--14903, 2016.

\bibitem{ref26}
S.~K. Ehrlich and G.~Cheng, ``A feasibility study for validating robot actions using eeg-based error-related potentials,'' {\em International Journal of Social Robotics}, vol.~11, pp.~271--283, 2019.

\bibitem{ref27}
S.~K. Ehrlich and G.~Cheng, ``Human-agent co-adaptation using error-related potentials,'' {\em Journal of neural engineering}, vol.~15, no.~6, p.~066014, 2018.

\bibitem{ref28}
F.~Iwane, I.~Iturrate, R.~Chavarriaga, and J.~del R~Mill{\'a}n, ``Invariability of eeg error-related potentials during continuous feedback protocols elicited by erroneous actions at predicted or unpredicted states,'' {\em Journal of Neural Engineering}, vol.~18, no.~4, p.~046044, 2021.

\bibitem{ref29}
C.~Lopes~Dias, A.~I. Sburlea, and G.~R. M{\"u}ller-Putz, ``Masked and unmasked error-related potentials during continuous control and feedback,'' {\em Journal of neural engineering}, vol.~15, no.~3, p.~036031, 2018.

\bibitem{ref30}
A.~F. Salazar-Gomez, J.~DelPreto, S.~Gil, F.~H. Guenther, and D.~Rus, ``Correcting robot mistakes in real time using eeg signals,'' in {\em 2017 IEEE international conference on robotics and automation (ICRA)}, pp.~6570--6577, IEEE, 2017.

\bibitem{ref31}
S.~Aldini, A.~K. Singh, M.~Carmichael, Y.-K. Wang, D.~Liu, and C.-T. Lin, ``Prediction-error negativity to assess singularity avoidance strategies in physical human-robot collaboration,'' in {\em 2021 IEEE International Conference on Robotics and Automation (ICRA)}, pp.~3241--3247, IEEE, 2021.

\bibitem{ref45}
M.~G. Carmichael, S.~Aldini, R.~Khonasty, A.~Tran, C.~Reeks, D.~Liu, K.~J. Waldron, and G.~Dissanayake, ``The anbot: An intelligent robotic co-worker for industrial abrasive blasting,'' in {\em 2019 IEEE/RSJ International Conference on Intelligent Robots and Systems (IROS)}, pp.~8026--8033, IEEE, 2019.

\bibitem{ref46}
R.~W. Homan, J.~Herman, and P.~Purdy, ``Cerebral location of international 10--20 system electrode placement,'' {\em Electroencephalography and clinical neurophysiology}, vol.~66, no.~4, pp.~376--382, 1987.

\bibitem{ref47}
C.~A.~E. Kothe and T.-P. Jung, ``Artifact removal techniques with signal reconstruction,'' Apr.~28 2016.
\newblock US Patent App. 14/895,440.

\bibitem{ref32}
S.~Makeig, A.~Bell, T.-P. Jung, and T.~J. Sejnowski, ``Independent component analysis of electroencephalographic data,'' {\em Advances in neural information processing systems}, vol.~8, 1995.

\bibitem{ref48}
A.~Mognon, J.~Jovicich, L.~Bruzzone, and M.~Buiatti, ``Adjust: An automatic eeg artifact detector based on the joint use of spatial and temporal features,'' {\em Psychophysiology}, vol.~48, no.~2, pp.~229--240, 2011.

\bibitem{ref50}
V.~Fran{\c{c}}ois-Lavet, P.~Henderson, R.~Islam, M.~G. Bellemare, J.~Pineau, {\em et~al.}, ``An introduction to deep reinforcement learning,'' {\em Foundations and Trends{\textregistered} in Machine Learning}, vol.~11, no.~3-4, pp.~219--354, 2018.

\bibitem{ref43}
T.~P. Lillicrap, J.~J. Hunt, A.~Pritzel, N.~Heess, T.~Erez, Y.~Tassa, D.~Silver, and D.~Wierstra, ``Continuous control with deep reinforcement learning,'' {\em arXiv preprint arXiv:1509.02971}, 2015.

\bibitem{ref53}
S.~Russell, ``Learning agents for uncertain environments,'' in {\em Proceedings of the eleventh annual conference on Computational learning theory}, pp.~101--103, 1998.

\bibitem{ref54}
R.~T. Schirrmeister, J.~T. Springenberg, L.~D.~J. Fiederer, M.~Glasstetter, K.~Eggensperger, M.~Tangermann, F.~Hutter, W.~Burgard, and T.~Ball, ``Deep learning with convolutional neural networks for eeg decoding and visualization,'' {\em Human brain mapping}, vol.~38, no.~11, pp.~5391--5420, 2017.

\bibitem{ref55}
S.~Aldini, A.~Akella, A.~K. Singh, Y.-K. Wang, M.~Carmichael, D.~Liu, and C.-T. Lin, ``Effect of mechanical resistance on cognitive conflict in physical human-robot collaboration,'' in {\em 2019 international conference on robotics and automation (ICRA)}, pp.~6137--6143, IEEE, 2019.

\bibitem{ref33}
P.~Berg and M.~Scherg, ``A fast method for forward computation of multiple-shell spherical head models,'' {\em Electroencephalography and clinical Neurophysiology}, vol.~90, no.~1, pp.~58--64, 1994.

\bibitem{ref34}
A.~Umemoto, M.~Inzlicht, and C.~B. Holroyd, ``Electrophysiological indices of anterior cingulate cortex function reveal changing levels of cognitive effort and reward valuation that sustain task performance,'' {\em Neuropsychologia}, vol.~123, pp.~67--76, 2019.

\bibitem{ref35}
M.~S. Erden and A.~Billard, ``End-point impedance measurements at human hand during interactive manual welding with robot,'' in {\em 2014 IEEE international conference on robotics and automation (ICRA)}, pp.~126--133, IEEE, 2014.

\bibitem{ref36}
J.~Harper, S.~M. Malone, and W.~G. Iacono, ``Theta-and delta-band eeg network dynamics during a novelty oddball task,'' {\em Psychophysiology}, vol.~54, no.~11, pp.~1590--1605, 2017.

\bibitem{ref38}
M.~M. Botvinick, T.~S. Braver, D.~M. Barch, C.~S. Carter, and J.~D. Cohen, ``Conflict monitoring and cognitive control.,'' {\em Psychological review}, vol.~108, no.~3, p.~624, 2001.

\bibitem{ref39}
L.~Gehrke, S.~Akman, P.~Lopes, A.~Chen, A.~K. Singh, H.-T. Chen, C.-T. Lin, and K.~Gramann, ``Detecting visuo-haptic mismatches in virtual reality using the prediction error negativity of event-related brain potentials,'' in {\em Proceedings of the 2019 CHI conference on human factors in computing systems}, pp.~1--11, 2019.

\bibitem{ref40}
T.~Ergenoglu, T.~Demiralp, Z.~Bayraktaroglu, M.~Ergen, H.~Beydagi, and Y.~Uresin, ``Alpha rhythm of the eeg modulates visual detection performance in humans,'' {\em Cognitive brain research}, vol.~20, no.~3, pp.~376--383, 2004.

\bibitem{ref41}
H.~T. Schupp, T.~Flaisch, J.~Stockburger, and M.~Jungh{\"o}fer, ``Emotion and attention: event-related brain potential studies,'' {\em Progress in brain research}, vol.~156, pp.~31--51, 2006.

\bibitem{ref42}
W.~J. Ray and H.~W. Cole, ``Eeg alpha activity reflects attentional demands, and beta activity reflects emotional and cognitive processes,'' {\em Science}, vol.~228, no.~4700, pp.~750--752, 1985.

\bibitem{ref44}
C.-T. Lin, C.-Y. Chiu, A.~K. Singh, J.-T. King, L.-W. Ko, Y.-C. Lu, and Y.-K. Wang, ``A wireless multifunctional ssvep-based brain--computer interface assistive system,'' {\em IEEE Transactions on Cognitive and Developmental Systems}, vol.~11, no.~3, pp.~375--383, 2018.

\end{thebibliography}

\end{document}